\documentclass[10pt,twocolumn,letterpaper]{article}

\usepackage{amsmath,amsfonts,bm}

\def\eqref#1{equation~\ref{#1}}
\def\1{\bm{1}}

\def\va{{\bm{a}}}

\def\vm{{\bm{m}}}

\def\vp{{\bm{p}}}

\def\vs{{\bm{s}}}

\def\vv{{\bm{v}}}

\def\vx{{\bm{x}}}

\def\vz{{\bm{z}}}

\def\mA{{\bm{A}}}

\DeclareMathAlphabet{\mathsfit}{\encodingdefault}{\sfdefault}{m}{sl}
\SetMathAlphabet{\mathsfit}{bold}{\encodingdefault}{\sfdefault}{bx}{n}

\usepackage{cvpr}              % To produce the CAMERA-READY version

\usepackage{graphicx,tabulary,amsmath,amssymb,booktabs,multirow,multicol,xspace,xcolor,bbding,comment,algorithm,listings,pifont,colortbl,bbold,makecell}
\usepackage[pagebackref,breaklinks,colorlinks]{hyperref}

\usepackage[capitalize]{cleveref}
\crefname{section}{Sec.}{Secs.}
\Crefname{section}{Section}{Sections}
\Crefname{table}{Table}{Tables}
\crefname{table}{Tab.}{Tabs.}

\def\x{$\times$}
\newcommand{\app}{\raise.17ex\hbox{$\scriptstyle\sim$}}
\newlength\savewidth\newcommand\shline{\noalign{\global\savewidth\arrayrulewidth
  \global\arrayrulewidth 1pt}\hline\noalign{\global\arrayrulewidth\savewidth}}
\newcommand{\tablestyle}[2]{\setlength{\tabcolsep}{#1}\renewcommand{\arraystretch}{#2}\centering\footnotesize}
\makeatletter\renewcommand\paragraph{\@startsection{paragraph}{4}{\z@}
  {.5em \@plus1ex \@minus.2ex}{-.5em}{\normalfont\normalsize\bfseries}}\makeatother

\newcolumntype{x}[1]{>{\centering\arraybackslash}p{#1pt}}
\newcolumntype{a}[1]{>{\columncolor{verylightgray}\centering\arraybackslash}p{#1pt}}
\newcolumntype{y}[1]{>{\raggedright\arraybackslash}p{#1pt}}
\newcolumntype{z}[1]{>{\raggedleft\arraybackslash}p{#1pt}}

\definecolor{baselinecolor}{gray}{.9}
\newcommand{\baseline}[1]{\cellcolor{baselinecolor}{#1}}

\def\cvprPaperID{3532} % *** Enter the CVPR Paper ID here
\def\confName{CVPR}
\def\confYear{2022}

\begin{document}

\title{\Large Point-Level Region Contrast for Object Detection Pre-Training\vspace{-.1em}}

\author{
Yutong Bai\textsuperscript{1,2}\thanks{Work done during an internship at FAIR.}\enspace \enspace
Xinlei Chen\textsuperscript{1}\enspace
Alexander Kirillov\textsuperscript{1}\enspace 
Alan Yuille\textsuperscript{2}\enspace
Alexander C. Berg\textsuperscript{1}\\
\vspace{.5em}
\textsuperscript{1}Facebook AI Research (FAIR) \qquad \qquad \textsuperscript{2}Johns Hopkins University %\vspace{.2em}
}
\maketitle

\begin{abstract}
\vspace{-.5em}
In this work we present point-level region contrast, a self-supervised pre-training approach for the task of object detection. This approach is motivated by the two key factors in detection: localization and recognition. While accurate localization favors models that operate at the pixel- or point-level, correct recognition typically relies on a more holistic, region-level view of objects. Incorporating this perspective in pre-training, our approach performs contrastive learning by directly sampling individual point pairs from different regions. Compared to an aggregated representation per region, our approach is more robust to the change in input region quality, and further enables us to implicitly improve initial region assignments via online knowledge distillation during training. Both advantages are important when dealing with imperfect regions encountered in the unsupervised setting. Experiments show point-level region contrast improves on state-of-the-art pre-training methods for object detection and segmentation across multiple tasks and datasets, and we provide extensive ablation studies and visualizations to aid understanding. Code will be made available. 
\vspace{-.5em}
\end{abstract}

\section{Introduction\label{sec:intro}}

Un-/self-supervised learning -- in particular contrastive learning~\cite{henaff2019data,moco,simclr} -- has recently arisen as a powerful tool to obtain visual representations that can potentially benefit from an unlimited amount of \emph{unlabeled} data. Promising signals are observed on important tasks like object detection~\cite{coco}. For example, MoCo~\cite{moco} shows convincing improvement on VOC~\cite{voc} over supervised pre-training by simply learning to discriminate between images as holistic instances~\cite{dosovitskiy2014discriminative} on the ImageNet-1K dataset~\cite{russakovsky2015imagenet}. Since then, numerous pre-text tasks that focus on \emph{intra-image} contrast have been devised specifically for object detection as the downstream transfer task~\cite{wang2020dense,xie2021propagate,henaff2021efficient}. While there has been steady progress, state-of-the-art detectors~\cite{lvis_challenge_site} still use weights from supervised pre-training (\eg, classification on ImageNet-22K~\cite{imagenet}). The full potential of unsupervised pre-training for object detection is yet to be realized.

%##################################################################################################
\begin{figure}[t]
  \centering
  \includegraphics[width=\linewidth]{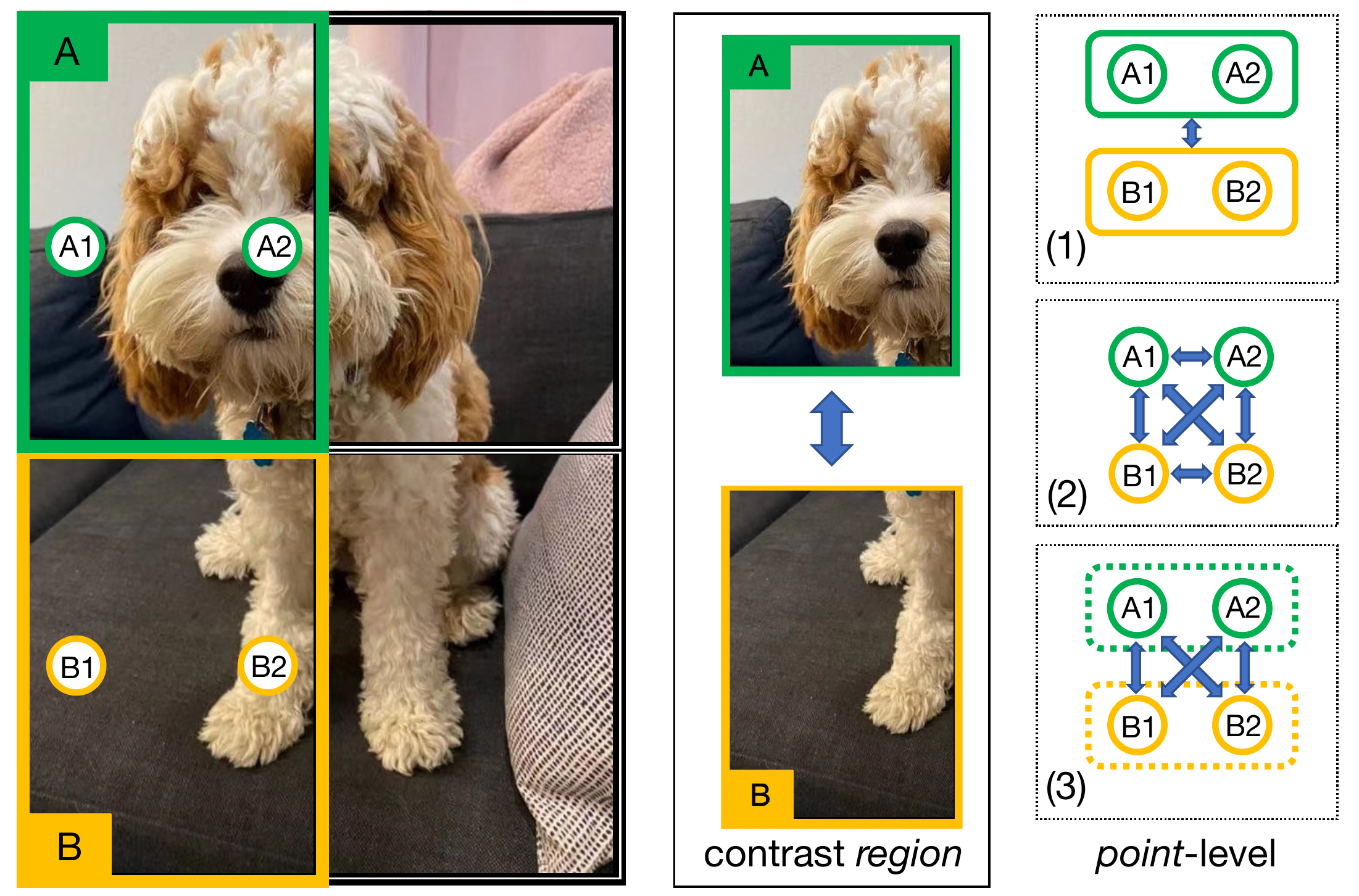}
  \caption{For intra-image contrastive learning, samples of a feature map can be aggregated and then compared between regions (1), compared directly between all samples (2), or only compared directly between samples in different regions (3).  We call (3) {\textbf{point-level region contrast}},  
it allows both learning at the point-level to help localization, and at the region-level to help holistic object recognition -- two crucial aspects for object detection.\label{fig:teaser}}
  \vspace{-.5em}
\end{figure}
%##################################################################################################

Object detection requires both accurate \emph{localization} of objects in an image and correct \emph{recognition} of their semantic categories. These two sub-tasks are tightly connected and often reinforce each other in successful detectors~\cite{malik2016three}. For example, region proposal methods~\cite{uijlings2013selective,zitnick2014edge,arbelaez2014multiscale} that first narrow down candidate object locations have enabled R-CNN~\cite{rcnn} to perform classification on rich, \emph{region-level} features. Conversely, today's dominant paradigm for object instance segmentation~\cite{he2017mask} first identifies object categories along with their coarse bounding boxes, and later uses them to compute masks for better localization at the \emph{pixel-level}.

With this perspective, we hypothesize that to learn a useful representation for object detection, it is also desirable to balance recognition and localization by leveraging information at various levels during pre-training. Object recognition in a scene typically takes place at the region-level~\cite{rcnn,ren2015faster}.  To support this, it is preferable to maintain a conceptually coherent `label' for each region, and learn to contrast pairs of regions 
%(rather than all pairs of pixels) 
for representation learning. On the other hand, for better localization, the model is preferred to operate at the pixel-, or \emph{`point-level'}~\cite{kirillov2020pointrend,cheng2021pointly}, especially when an initial, unsupervised, assignment of pixels to regions (\ie, segmentation) is sub-optimal (see \cref{fig:teaser} for an example). To our knowledge, existing methods in this frontier can be lacking in either of these two aspects (to be discussed in \cref{sec:related}).

In this paper, we present a self-supervised pre-training approach that conceptually contrasts at the region-level while operating at the point-level. Starting from MoCo v2~\cite{chen2020improved} as an image-level baseline, we introduce the notion of `regions' by dividing each image into a non-overlapping grid~\cite{henaff2021efficient}. Treating rectangular regions on this grid as separate instances, we can define the task of intra-image discrimination on top of the existing inter-image one~\cite{dosovitskiy2014discriminative} and pre-train a representation with contrastive objectives. Deviating from the common practice that aggregates features for contrastive learning~\cite{moco,simclr,henaff2021efficient}, we directly operate at the point-level by sampling multiple points from each region, and contrasting point pairs individually across regions (see \cref{fig:teaser}, right column for illustrations).

The advantage of operating at the point-level is two-fold, both concerning dealing with \emph{imperfect} regions as there is no ground-truth. First, such a design can be more \emph{robust} to the change in region quality, since feature aggregation can cause ambiguities when the regions are not well localized (\eg, in \cref{fig:teaser}, both regions of interest can mean `a mixture of dog and couch'), whereas individual points still allow the model to see distinctions. 
Second and perhaps more importantly, it can enable us to \emph{bootstrap}~\cite{grill2020bootstrap} for potentially better regions during the training process. This is because any segmentation can be viewed as a hard-coded form of \emph{point affinities} -- 1 for point pairs within the same region and 0 otherwise; and a natural by-product of contrasting point pairs is soft point affinities (values between 0 and 1) that \emph{implicitly} encode regions. By viewing the momentum encoder as a `teacher' network, we can formulate the problem as knowledge distillation one~\cite{hinton2015distilling,caron2021emerging}, and improving point affinities (and thus implicitly regions) online in the same self-supervised fashion. 

Empirically, we applied our approach to standard pre-training datasets (ImageNet-1K~\cite{imagenet} and COCO train set~\cite{coco}), and transferred the representation to multiple downstream datasets: VOC~\cite{voc}, COCO (for both object detection and instance segmentation), and Cityscapes~\cite{cityscapes} (semantic segmentation). We show strong results compared to state-of-the-art pre-training methods which use image-level, point-level, or region-level contrastive learning. Moreover, we provide extensive ablation studies covering different aspects in design, and qualitatively visualize the point affinities learned through knowledge distillation. % \xl{iterate on this paragraph needed}

While we are yet to showcase improvements on larger models, longer training schedules, stronger augmentations~\cite{ghiasi2021simple}, and bigger pre-training data for object detection, we believe our explorations on the pre-training design that better balances recognition and localization can inspire more works in this direction.

\section{Related Work\label{sec:related}}

\paragraph{Self-supervised learning.} Supervised learning/classification~\cite{he2016deep,russakovsky2015imagenet} has been the dominant method for pre-training representations useful for downstream tasks in computer vision. Recently, contrastive learning~\cite{wu2018unsupervised,henaff2019data,tian2019contrastive,moco,simclr,dwibedi2021little} has emerged as a promising alternative that pre-trains visual representations \emph{without} class labels or other forms of human annotations -- a paradigm commonly referred as `self-supervised learning'. By definition, self-supervised learning holds the potential of scaling up pre-training to huge models and billion-scale data. As a demonstration, revolutionary progress has already been made in fields like natural language processing~\cite{devlin2018bert,Radford2019,Brown2020} through scaling. For computer vision, such a moment is yet to happen. Nonetheless, object detection as a fundamental task in computer vision is a must-have benchmark to test the transferability of pre-trained representations~\cite{rcnn}.

\paragraph{Contrastive learning.} Akin to supervised learning which maps images to class labels, contrastive learning maps images to separate vector embeddings, and attracts positive embedding pairs while dispels negative pairs. A key concept connecting the two types of learning is instance discrimination~\cite{dosovitskiy2014discriminative}, which models each image as its \emph{own} class. Under this formulation, two augmentations of the same image is considered as a positive pair, while different images form negative pairs. Interestingly, recent works show that negative pairs are not required to learn meaningful representations~\cite{grill2020bootstrap,chen2021exploring} for reasons are yet to be understood. Regardless, all these frameworks treat each image as a single instance and use aggregated (\ie, pooled) features to compute embeddings. Such a classification-oriented design largely ignores the internal structures of images, which could limit their application to object detection that performs dense search \emph{within} an image~\cite{ren2015faster,liu2016ssd,lin2017focal}.

\paragraph{Point-level contrast.} Many recent works~\cite{wang2020dense,xie2021propagate,liu2020self,xie2020pointcontrast,pinheiro2020unsupervised} have realized the above limitation, and extended the original idea from contrasting features between whole images to contrasting features at points. Different ways to match points as pairs have been explored. For example, \cite{wang2020dense} selects positive pairs by ranking similarities among all points in the latent space; \cite{xie2021propagate} defines positive pairs by spatial proximity; \cite{liu2020self} jointly matches a set of features at points to another set via Sinkhorn-Knopp algorithm~\cite{cuturi2013sinkhorn}, designed to maximize the set-level similarity for sampled features. However, we believe directly contrasting features at arbitrary points over-weights localization, and as a result misses a more global view of the entire object that can lead to better \emph{recognition}.

\paragraph{Region-level contrast.} Closest to our paper is the most recent line of work that contrasts representations at the region-level~\cite{henaff2021efficient,xie2021detco,wei2021aligning,xie2021unsupervised,xiao2021region,roh2021spatially}. Specifically, images are divided into regions of interest, via either external input~\cite{henaff2021efficient,wei2021aligning,xie2021unsupervised}, or sliding windows~\cite{xiao2021region}, or just random sampling~\cite{xie2021detco,roh2021spatially}. Influenced by image-level contrastive learning, most approaches represent each region with a single, aggregated vector embedding for loss computation and other operations, which we argue -- and show empirically -- is detrimental for \emph{localization} of objects.

%##################################################################################################
\begin{figure*}[t]
    \centering
    \includegraphics[width=0.8\linewidth]{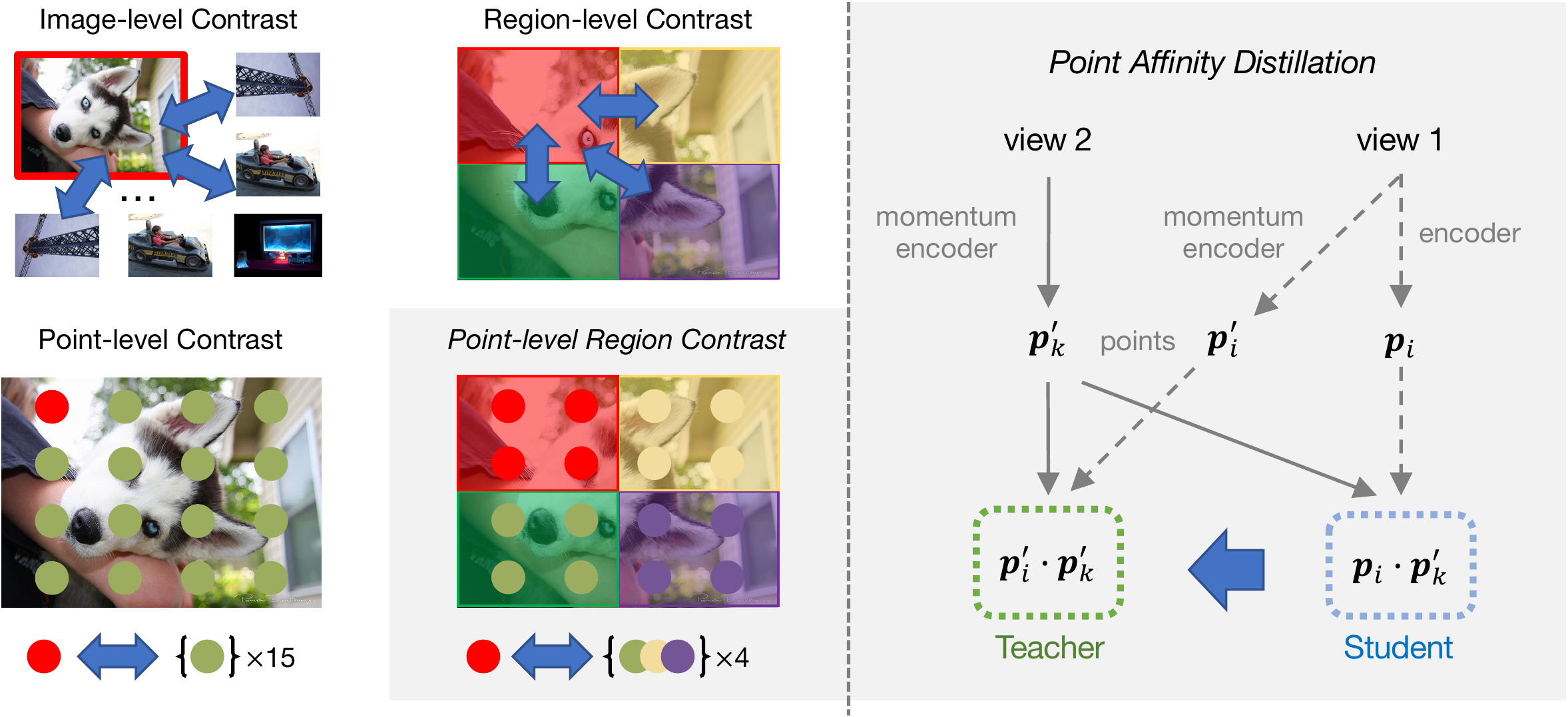}
    \vspace{-.5em}
    \caption{\label{fig:method}Illustration of \textbf{point-level region contrast} (\cref{sec:contrast}), which also enables \textbf{point affinity distillation} (\cref{sec:knowledge}). On the left we show four different types of contrastive learning methods, including image-level, region-level, point-level and our point-level region contrast. On the right we show point affinity distillation with one pair of points.}
    \vspace{-.5em}
\end{figure*}
%##################################################################################################

\section{Approach\label{sec:approach}}

In this section we detail our approach: point-level region contrast. To lay the background and introduce notations, we begin by reviewing the formulation of MoCo~\cite{moco}.

\subsection{Background: Momentum Contrast\label{sec:background}}

As the name indicates, MoCo~\cite{moco,chen2020improved} is a contrastive learning framework~\cite{oord2018representation,simclr} that effectively uses momentum encoders to learn representations. Treating each image as a single \emph{instance} to discriminate against others, MoCo operates at the image-level (see \cref{fig:method} top left corner).

\paragraph{Image-level contrast.} While the original model for instance discrimination~\cite{dosovitskiy2014discriminative} literally keeps a dedicated weight vector for each image in the dataset (on ImageNet-1K~\cite{russakovsky2015imagenet} it would mean more than one million vectors), modern frameworks~\cite{wu2018unsupervised,simclr} formulate this task as a contrastive learning one which only requires online computation of embedding vectors per-image and saves memory. Specifically MoCo, two parallel encoders, $f^{E}$ and $f^{M}$, take two augmented views ($\vv$ and $\vv'$) for each image $\vx$ in a batch, and output two $\ell_2$-normalized embeddings $\vz$ and $\vz'$. Here $f^E$ denotes the base encoder being trained by gradient updates as in normal supervised learning, and $f^M$ denotes the momentum encoder that keeps updated by exponential moving average on the base encoder weights. Then image-level contrastive learning is performed by enforcing similarity on views from the same image, and dissimilarity on views from different images, with the commonly used InfoNCE objective~\cite{oord2018representation}:
%##################################################################################################
\begin{equation}
\small
\mathcal{L}_\text{m} = -\log \frac{\exp(\vz{\cdot}\vz' / \tau)}{\sum_{j}\exp(\vz{\cdot}\vz'_j / \tau)},
\label{eq:infonce}
\end{equation}
%##################################################################################################
Where $\tau$ is the temperature, other images (and self) are indexed by $j$. In MoCo, other images are from the momentum bank~\cite{wu2018unsupervised}, which is typically much smaller in size compared to the full dataset. 

It is important to note that in order to compute the embedding vectors $\vz$ (and $\vz'$), a pooling-like operation is often used in intermediate layers to aggregate information from all spatial locations in the 2D image. This is inherited from the practice in supervised learning, where standard backbones (\eg, ResNet-50~\cite{he2016deep}) average-pool features before the classification task.

\subsection{Point-Level Region Contrast\label{sec:contrast}}

As discussed above, image-level contrast is classification oriented. Next, we discuss our designs in point-level region contrast, which are more fit for the tasks of object detection.

\paragraph{Regions.} Region is a key concept in state-of-the-art object detectors~\cite{ren2015faster,he2017mask}. Through region-of-interest pooling, object-level recognition (\ie, classifying objects into pre-defined categories) are driven by region-level features. Different from detector training, ground-truth object annotations are not accessible in self-supervised learning. Therefore, we simply introduce the notion of regions by dividing each image into a non-overlapping, $n$\x$n$ grid~\cite{henaff2021efficient}. We treat the rectangular regions on this grid as separate instances, which allows inter-image contrast and \emph{intra-image} contrast to be jointly performed on pairs of regions. Now, each augmentation $\vv$ is paired with masks, and each mask denotes the corresponding region under the same geometric transformation as $\vv$ with which it shares resolution. Note that due to randomly resized cropping~\cite{moco}, some masks can be empty. Therefore, we randomly sample $N{=}16$ valid masks $\{\vm_n\}$ ($n{\in}\{1,\ldots,N\}$) (with repetition) as regions to contrast, following the design of~\cite{henaff2021efficient}.

Grid regions are the simplest form of the spatial heuristic that nearby pixels are likely belong to the same object~\cite{henaff2021efficient}. More advanced regions~\cite{uijlings2013selective,arbelaez2014multiscale}, or even ground-truth segmentation masks (used for analysis-only)~\cite{coco} can be readily plugged in our method to potentially help performance, but it comes at the expense of more computation costs, potential risk of bias~\cite{chavali2016object} or human annotation costs. Instead, we focus on improving training strategies and just use grids for our explorations.

\paragraph{Point-level.} Given the imperfect regions, our key insight is to operate at the point-level. Intuitively, pre-training by contrasting regions can help learn features that are discriminative enough to tell objects apart as \emph{holistic} entities, but they can be lacking in providing low-level cues for the exact \emph{locations} of objects. This is particularly true if features that represent regions are aggregated over all pertinent locations, just like the practice in image-level contrast. Deviating from this, we directly sample multiple points from each region, and contrast point pairs individually across regions \emph{without} pooling.

Formally, we sample $P$ points per mask $\vm_n$, and compute point-level features ${\vp_i}$ ($i{\in}\{1,\ldots,N{\times}P\}$) for contrastive learning. Each $\vp_i$ comes with an indicator for its corresponding region, $\va_i$. To accommodate this, we modify the encoder architecture so that the \emph{spatial} dimensions are kept all the way till the output.\footnote{An additional projector MLP is introduced in MoCo v2~\cite{chen2020improved} following SimCLR~\cite{simclr}, we convert the MLP into 1\x1 convolution layers.} The final feature map is up-sampled to a spatial resolution of $R{\times}R$ via interpolation. Then our point-level, region contrastive loss is defined as:
%##################################################################################################
\begin{equation}
\small
\mathcal{L}_\text{c} = -\frac{1}{C}\sum_{\va_i{=}\va_k} {\log \frac{\exp(\vp_i{\cdot}\vp'_k / \tau)}{\sum_{j}\exp(\vp_i{\cdot}\vp'_j / \tau)}},
\label{eq:infpoint}
\end{equation}
%##################################################################################################
where $j$ loops over points from regions in the same image (intra-), or over points from other images (inter-). $C$ is a normalization factor for the loss which depends on the number of positive point pairs. An illustrative case (for $n{=}2$ and $P{=}4$) is shown in \cref{fig:method}.

\subsection{Point Affinity Distillation\label{sec:knowledge}}

Operating at the point level enables us to bootstrap~\cite{zhang2020self} and not be restricted by the pre-defined regions. This is because according to \cref{eq:infpoint}, the only place the pre-defined regions matter is in the indicators $\va_i$, which provides a \emph{hard} assignment from points to regions. When $\va_i{=}\va_k$, it means the probability of $\vp_i$ and $\vp_k$ coming from the same region is 1, otherwise it is 0.

On the other hand, the InfoNCE loss~\cite{oord2018representation} (\cref{eq:infonce}) used for contrastive learning computes \emph{point affinities} as a natural by-product which we define as:
\begin{equation}
\small
\mA_{ik'}(\tau) := \frac{\exp(\vp_i{\cdot}\vp'_k / \tau)}{\sum_{j}\exp(\vp_i{\cdot}\vp'_j / \tau)}.
\label{eq:affinity}
\end{equation}
Note that $\mA_{ik'}(\tau)$ is a pairwise term controlled by two indexes $i$ and $k'$, and the additional $'$ indicates the embeddings participating is computed by the momentum encoder. For example $\mA_{i'k'}(\tau)$ means both embeddings are from the momentum encoder $f^M$. Point affinities offer \emph{soft}, implicit assignment from points to regions, and an explicit assignment can be obtained via clustering (\eg k-means). In this sense, they arguably provide more \emph{complete} information about which point pairs belong to the same region.

The Siamese architecture~\cite{chen2021exploring} of self-supervised learning methods like MoCo presents a straightforward way to bootstrap and obtain potentially better regions. The momentum encoder $f^M$ itself can be viewed as a `teacher' which serves as a judge for the quality of $f^E$~\cite{caron2021emerging}. From such an angle, we can formulate the problem as a knowledge distillation one~\cite{hinton2015distilling}, and use the outputs of $f^M$ to supervise the point affinities that involve $f^E$ via cross entropy loss:
%##################################################################################################
\begin{equation}
\small
\mathcal{L}_\text{a} = - \sum_{i,k} \mA_{i'k'}(\tau_t) \log \mA_{ik'}(\tau_s),
\label{eq:kd}                                                                
\end{equation}
%##################################################################################################
where $\tau_t$ and $\tau_s$ are temperatures for the teacher and the student, respectively. We call this `point affinity distillation'.
There are other possible ways to distill point affinities from the momentum encoder (see~\cref{sec:ablatedistillation}), we choose the current design trading off speed and accuracy.

On the other hand, we note that the pooling operation does \emph{not} back-propagate gradients to the coordinates (only to the features) by default. Therefore, it is less straightforward to morph regions along with training by contrasting aggregated region-level features~\cite{henaff2021efficient,wei2021aligning,xie2021unsupervised}.

%##################################################################################################
\begin{table*}[t]
\centering\small
\tablestyle{2pt}{1.1}
\begin{tabular}{y{50}|x{30}|x{30}x{30}x{30}|x{30}x{30}x{30}|x{30}x{30}x{30}|x{40}}
\multirow{2}{*}{method} & \multirow{2}{*}{\makecell{\# of \\epochs}} & \multicolumn{3}{c|}{Pascal VOC}  & \multicolumn{3}{c|}{COCO detection} & \multicolumn{3}{c|}{COCO segmentation} & Cityscapes \\
 & & AP & $\text{AP}_\text{50}$ & $\text{AP}_\text{75}$ & AP & $\text{AP}_\text{50}$ & $\text{AP}_\text{75}$ & AP & $\text{AP}_\text{50}$ & $\text{AP}_\text{75}$ & mIoU\\
\shline
Scratch & - & 33.8 & 60.2 & 33.1 & 26.4 & 44.0 & 27.8 & 29.3 & 46.9 & 30.8  & 65.3 \\
\hline
Supervised & 200 & 54.2 & 81.6 & 59.8 & 38.2 & 58.2 & 41.2 & 33.3 & 54.7 & 35.2 &73.7  \\
\hline
MoCo~\cite{moco} & 200 & 55.9 & 81.5  & 62.6 & 38.5 & 58.3 & 41.6 & 33.6 & 54.8 & 35.6  & 75.3 \\
SimCLR~\cite{simclr} & 1000 & 56.3 & 81.9 & 62.5 & 38.4 & 58.3 & 41.6 & - & - & -  & 75.8 \\
MoCo v2~\cite{chen2020improved} & 800 & 57.6 & 82.7 & 64.4 & 39.8 & 59.8 & 43.6  & 36.1 & 56.9 & 38.7 & 76.2 \\
InfoMin~\cite{tian2020makes} & 200 & 57.6 & 82.7 & 64.6 & 39.0 & 58.5 & 42.0 & - & - & -  & 75.6 \\
% InfoMin~\cite{tian2020makes} & 800 & 57.5 & 82.5 & 64.0 & - & -  & - & 38.8 & 58.2 & 41.7 & 75.6 \\
\hline
DetCo~\cite{xie2021detco} & 200 & 57.8 & 82.6 & 64.2 & 39.8 & 59.7& 43.0 & 34.7 & 56.3 & 36.7 & 76.5\\
InsLoc~\cite{yang2021insloc} & 800 & 58.4 & 83.0 & 65.3 &39.8 &59.6 &42.9 &34.7 &56.3 &36.9 & - \\
% DenseCL~\cite{wang2020dense} & 200 & 58.7 & 82.8 & 65.2& 40.3 & 59.9 & 44.3  &36.4 & 57.0 & 39.2 & 75.7 \\
PixPro~\cite{xie2021propagate} & 200 & 58.8 & 83.0 & 66.5 & 40.0 & 59.3 & 43.4 & 34.8 & - & -  & 76.8\\
% \hline
DetCon~\cite{henaff2021efficient} & 200 & -  & -  & -  &  40.5  & - & - & 36.4  & - & - & 76.5  \\
SoCo~\cite{wei2021aligning} & 200 & 59.1 & 83.4 & 65.6  & 40.4 & \textbf{60.4} & 43.7 & 34.9 & 56.8 & 37.0 & 76.5  \\
\hline
\emph{Ours}& 200 & \textbf{59.4} &  \textbf{83.6}&  \textbf{67.1} & \textbf{40.7}& \textbf{60.4}& \textbf{44.7} &  \textbf{36.9} & \textbf{57.4} & \textbf{39.6} & \textbf{77.0} \\
\end{tabular}
\vspace{-.2em}
\caption{\textbf{Main results with ImageNet-1K pre-training.} From left to right, we show transfer performance on 4 tasks: VOC (07+12) detection~\cite{voc}, COCO object detection~\cite{coco}; COCO instance segmentation and Cityscapes semantic segmentation~\cite{cityscapes}. From top to down, we compare our approach with 3 other setups: i) no pre-training (\ie, scratch); ii) general pre-training with supervised learning or \emph{inter-image} contrastive learning; iii) object detection oriented pre-training with additional \emph{intra-image} contrast. Our point-level region contrast pre-training shows consistent improvements across different tasks under fair comparisons. \label{tab:main}}
\vspace{-.5em}
\end{table*}
%##################################################################################################

\subsection{Overall Loss Function\label{sec:overall}}

We jointly perform point-level region contrast learning (\cref{sec:contrast}) and point affinity distillation (\cref{sec:knowledge}) controlled by a balance factor $\alpha$:
%##################################################################################################
\begin{equation}
    \small
    \mathcal{L_\text{p}} = \alpha \mathcal{L}_\text{c} + (1-\alpha) \mathcal{L}_\text{a}.
    \label{eq:point}
\end{equation}
Here, $\mathcal{L}_\text{c}$ offers an initialization of regions to contrast with, whereas $\mathcal{L}_\text{a}$ bootstraps~\cite{grill2020bootstrap} from data, regularizes learning and alleviates over-fitting to the initial imperfect region assignments. This is how these two terms interact and benefit each other -- a common practice for knowledge distillation with additional ground-truth labels. 

%##################################################################################################
Finally, our point-level loss is added to the original MoCo loss for joint optimization, controlled by another factor $\beta$:
%##################################################################################################
\begin{equation}
    \small
    \mathcal{L} = \beta \mathcal{L}_\text{p} + (1-\beta) \mathcal{L}_\text{m},
    \label{eq:total_loss}
\end{equation}
%##################################################################################################
which does not incur extra overhead for backbone feature computation. Note that all the loss terms we have defined above are focused on a single image for explanation clarity, the full loss is averaged over all images.

\section{Experiments\label{sec:experiments}}

In this section we perform experiments. For our main results, we pre-train on ImageNet-1K or COCO, and transfer the learned representations to 4 downstream tasks. 
We then conduct analysis by: 1) visualizing the learned point affinities with a quantitative evaluation metric using VOC ground-truth masks, 2) presenting evidence that point-level representations are effective and more robust to region-level ones when the mask quality degenerates; and 3) ablating different point affinity distillation strategies in our approach. 
More analysis on various hyper-parameters and more visualizations are found in the appendix.

\subsection{Pre-Training Details\label{sec:exp_setting}}
We either pre-train on ImageNet-1K~\cite{russakovsky2015imagenet} or COCO~\cite{coco}, following standard setups~\cite{henaff2021efficient,wang2020dense}. 

\paragraph{ImageNet-1K setting.}
Only images from the training split are used, which leads to \app1.28 million images for ImageNet-1K. We pre-train the model for 200 epochs. 

It is worth noting that we build our approach on the default, \emph{asymmetric} version of MoCo v2~\cite{chen2020improved}, which is shown to roughly compensate for the performance of pre-training with \emph{half} the length using \emph{symmetrized} loss~\cite{chen2021exploring} -- both setups share the same amount of compute in this case. 

\paragraph{COCO setting.}
Only images from the training split ({\fontfamily{pcr}\selectfont train2017}) are used, which leads to \app118k for COCO. We pre-train with 800 \emph{COCO} epochs, not ImageNet epochs.

\paragraph{Hyper-parameters and augmentations.} We use a 4\x4 grid and sample $N{=}16$ valid masks per view following~\cite{henaff2021efficient}. $P{=}16$ points are sampled per region. The up-sampled resolution of the feature map $R$ is set to $64$. We use a teacher temperature $\tau_t$ of 0.07 and student temperature $\tau_s$ of 0.1, with 30 epochs as a warm-up stage where no distillation is applied. The balancing ratios for losses are set as $\alpha{=}0.5$ and $\beta{=}0.7$. For optimization hyper-parameters (\eg learning rate, batch size \etc) and augmentation recipes we follow MoCo v2~\cite{chen2020improved}. We follow the same strategy in DetCon~\cite{henaff2021efficient} to sample region pairs through random crops, and skip the loss computation for points when views share no overlapping region, which happens rarely in practice.

%##################################################################################################
\begin{table*}[th]
\centering\small
\tablestyle{2pt}{1.1}
\begin{tabular}{y{50}|x{30}|x{30}x{30}x{30}|x{30}x{30}x{30}|x{30}x{30}x{30}|x{40}}
\multirow{2}{*}{method} & \multirow{2}{*}{\makecell{\# of \\epochs}} & \multicolumn{3}{c|}{Pascal VOC} & \multicolumn{3}{c|}{COCO detection } & \multicolumn{3}{c|}{COCO segmentation} & Cityscapes \\
 & & AP & $\text{AP}_\text{50}$ & $\text{AP}_\text{75}$ & AP & $\text{AP}_\text{50}$ & $\text{AP}_\text{75}$ & AP & $\text{AP}_\text{50}$ & $\text{AP}_\text{75}$ & mIoU\\
\shline
Scratch & - & 33.8 & 60.2 & 33.1 & 29.9 & 47.9 & 32.0 & 32.8 & 50.9 & 35.3 & 63.5\\
\hline
% Supervised & - & 35.9 &56.6& 38.6 & \\
MoCo v2~\cite{chen2020improved} & 800 &   54.7 & 81.0 & 60.6 & 38.5 & 58.1 & 42.1 & 34.8 & 55.3 & 37.3  & 73.8 \\
BYOL~\cite{grill2020bootstrap} & 800 & -&-&- & 37.9 & 57.5& 40.9 & -&-&- &- \\
\hline
Self-EMD~\cite{liu2020self} & 800 & -  &  - &  - & 38.5 & 58.3 &41.6 & -&-&-  & -\\
PixPro~\cite{xie2021propagate} & 800 & 56.5   & 81.4 & 62.7 & 39.0 & 58.9 & 43.0 & 35.4 & 56.2  & 38.1 & 75.2 \\
% DenseCL~\cite{wang2020dense} & 800 &   56.7 & 81.7 & 63.0 & 39.6 & 59.3 & 43.3 & 35.7 & 56.5 & 38.4 & 75.6 \\
% DetCon~\cite{wang2020dense} & 800 &    &  &  &  &  &  &  &  &  &  \\
\hline
\emph{Ours}& 800 & \textbf{57.1} & \textbf{82.1} & \textbf{63.8}  &\textbf{39.8}&\textbf{59.6}&\textbf{43.7}  & \textbf{35.9}&\textbf{56.9}&\textbf{38.6} & \textbf{75.9} \\
\end{tabular}
\vspace{-.2em}
\caption{\textbf{Main results with COCO pre-training.} Same as ImageNet-1K, from left to right, we show the performance on 4 tasks: VOC (07+12) detection, COCO detection; COCO instance segmentation and Cityscapes semantic segmentation. From top to down, we compare with training from scratch and pre-training with self-supervision. For COCO pre-training, our method shows significant improvements.
\label{tab:maincoco}}
\vspace{-.5em}
\end{table*}
%##################################################################################################

\subsection{Downstream Tasks\label{sec:downstream}}
We evaluate feature transfer performance on four downstream tasks: object detection on VOC~\cite{voc}, object detection and instance segmentation on COCO~\cite{coco}, and semantic segmentation on Cityscapes~\cite{cityscapes}. 

\paragraph{VOC.} PASCL VOC is the default dataset to evaluate self-supervised pre-training for object detection. We follow the setting introduced in MoCo~\cite{moco}, namely a Faster R-CNN detector~\cite{ren2015faster} with the ResNet-50 \emph{C4} backbone, which uses the \emph{conv4} feature map to produce object proposals and uses the \emph{conv5} stage for proposal classification and bounding box regression. In fine-tuning, we synchronize all batch normalization layers across devices. Training is performed on the combined set of {\fontfamily{pcr}\selectfont trainval2007} and {\fontfamily{pcr}\selectfont trainval2012}. For testing, we report AP, AP50 and AP75 on the {\fontfamily{pcr}\selectfont test2007} set. Detectron2~\cite{wu2019detectron2} is used.

\paragraph{COCO.} On COCO we study both object bounding box detection and instance segmentation. We adopt Mask R-CNN~\cite{he2017mask} with ResNet-50 \emph{C4} as the backbone and head. Other setups are the same as VOC. Detectron2 is again used. We follow the standard $1\times$ schedule for fine-tuning, which is 90k iterations for COCO.

\paragraph{Cityscapes.} On Cityscapes we evaluate semantic segmentation, a task that also relies on good localization and recognition.
We follow the previous settings~\cite{moco,xie2021propagate}, where a FCN-based structure is used~\cite{long2015fully}. 
The classification is obtained by an additional $1{\times}1$ convolutional layer.

\subsection{Main Results\label{sec:main_results}}

\paragraph{ImageNet-1K pre-training.} \cref{tab:main} compares our point-level region contrast to previous state-of-the-art unsupervised pre-training approaches on 4 downstream tasks, which all require dense predictions. We compare with four categories of methods: 1) training from scratch, \ie learning the network from random initialization; 2) ImageNet-1K supervised pre-training; 3) general self-supervised pre-training, including MoCo, MoCo v2, SimCLR and InfoMin. Those are under their reported epochs; 4) Task-specific pre-training, including DetCo~\cite{xie2021detco}, PixPro~\cite{xie2021propagate}, DenseCL~\cite{wang2020dense} and DetCon \cite{henaff2021efficient}. We report the numbers with 200-epoch pre-training. It is worth noting that we adopt the asymmetric network structure~\cite{chen2020improved}, \ie each view is only used once per iteration. For this reason, we denote PixPro (100-epoch reported in~\cite{wang2020dense}) and SoCo \cite{wei2021aligning} as 200 epochs since the loss is symmetrized there.
DetCon \cite{henaff2021efficient} uses pre-defined segmentation masks acquired by off-the-shelf algorithms. We also compared with it under the same number of epochs.

It shows consistent improvement on every tasks compared with prior arts under this \emph{fair} comparison setting on VOC object detection, COCO object detection, COCO instance segmentation and Cityscapes semantic segmentation. 

\paragraph{COCO pre-training.} \cref{tab:maincoco} compares our method to previous state-of-the-art unsupervised pre-training approaches on COCO.
We evaluate the transferring ability to the same 4 downstream tasks used for ImageNet-1K pre-training, and on all of them we show significant improvements. Different from ImageNet-1K, COCO images have more objects per-image on average, thus our point-level region contrast is potentially more reasonable and beneficial in this setting.

%##################################################################################################
\begin{figure*}[!t]
\centering
\includegraphics[width=\linewidth]{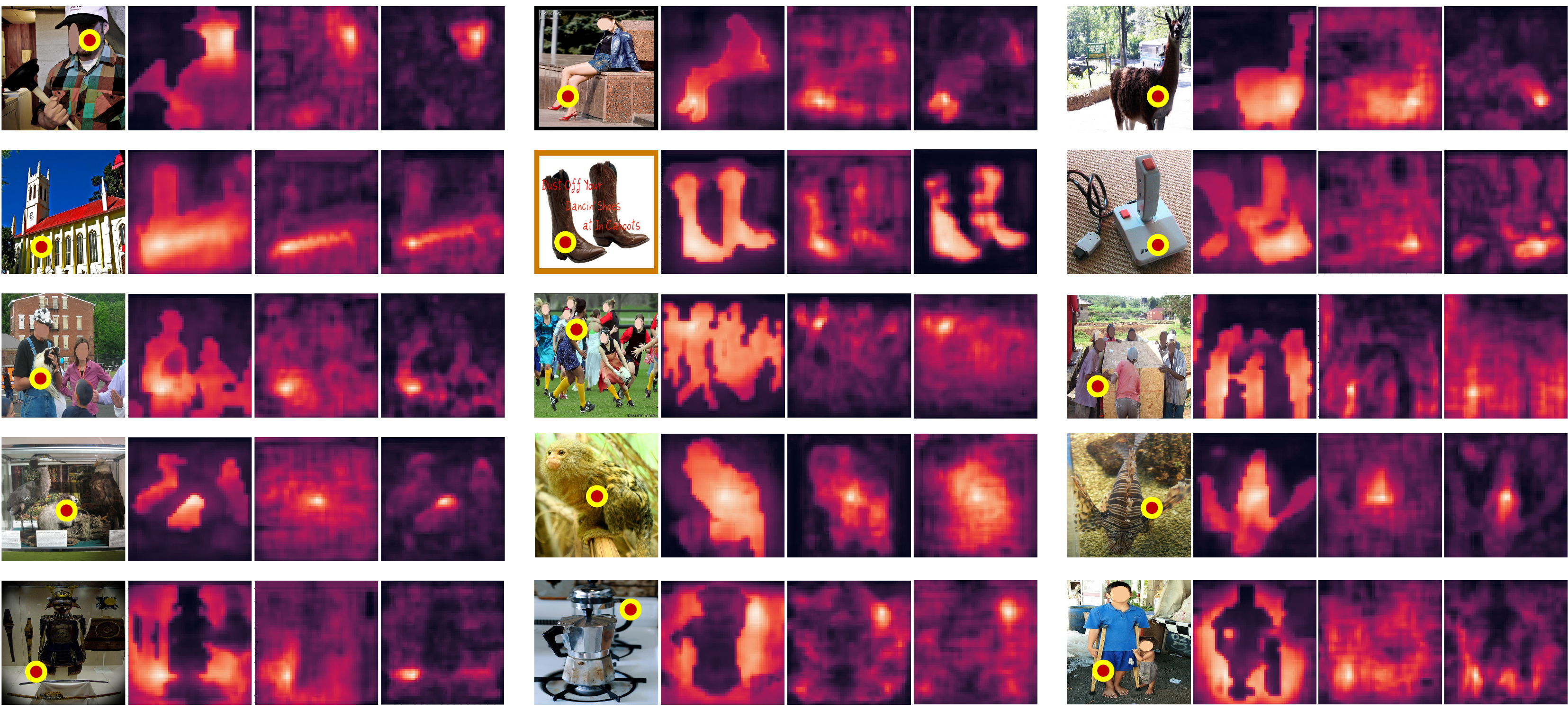}
\vspace{-1.5em}
\caption{\textbf{Point affinity visualizations.}
In total we show 15 groups of examples. In each group from left to right we show the original image with the selected point (denoted by red circle); three affinity maps calculated from the point to the rest of the image with the output of i) our point-level region contrast; ii) region-level contrast; and iii) MoCo v2 (image-level contrast). In rows from top to down, we show 5 categories of picked points: i) single non-rigid objects, ii) single rigid objects, iii) multiple objects, iv) objects in chaotic background and v) background stuff. Brighter colors in the affinity map denote more similar points. Best viewed in color and zoomed in.\label{fig:visualization}}
\vspace{-.5em}
\end{figure*}
%##################################################################################################

%##################################################################################################
\begin{figure*}[ht]
\centering
\includegraphics[width=\linewidth]{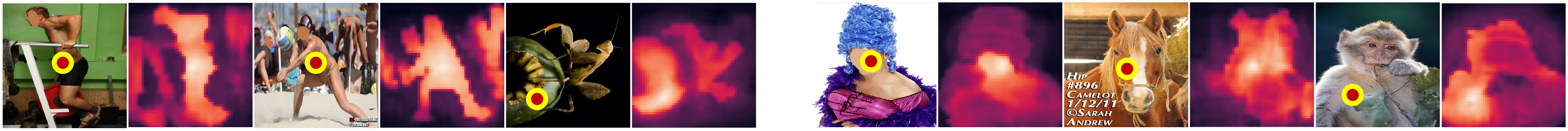}
\vspace{-1.5em}
\caption{\textbf{Point affinity (failures).} We present two kinds of failure cases for our method: under-segmentation (left) and over-segmentation (right). For each kind we show 3 pairs of images, using the same visualization technique in \cref{fig:visualization}. See text for details.\label{fig:badcase}}
\vspace{-1em}
\end{figure*}
%##################################################################################################

\subsection{Visualization of Point Affinities\label{sec:vis_point_affinity}}
In order to provide a more intuitive way to show the effectiveness of our method, we visualize the point affinities after pre-training in \cref{fig:visualization}.

The images are randomly chosen from the validation set of ImageNet-1K. We follow the previous experimental setting to pre-train 200 epochs on ImageNet-1K. We then resize all image to 896\x896, and interpolate the corresponding feature map of \emph{Res5} from (28\x28) to 56\x56 for higher resolution. 
For each image, we first pick one point (denoted with a red circle), then calculate the point affinity (in terms of cosine similarity) from the last-layer output feature representation of this point to all the others within the same image. 
In addition, we also compare it with the visualizations from MoCo v2 and a region-level contrast variant of our method to analyse the improvement.
The region-level contrast variant is implemented using MoCo v2 framework with grid regions (same as ours), with an AP of 58.2 on VOC.
In \cref{fig:visualization}, from top to down we show 15 different groups of examples which (row-wise) represent 5 categories of picked points: single non-rigid objects, single rigid objects, multiple objects, objects in chaotic background, and background. Within each group, from left to right we show the point affinity of our method, region-level contrast, and the MoCo v2 baseline. Brighter colors on the feature map denote more similar points. 

\paragraph{Observations.} For original MoCo, its final global pooling operation intuitively causes a loss in 2D spatial information, since everything is compressed into a single vector for representations. Therefore, when tracing back, the salient regions usually only cover certain closely-connected small area around the picked point.
For the region-level contrast baseline, its salient regions can expand to a larger area, but the area is quite blurry and hard to tell the boundaries.
For objects (shown in row 1-3), although all three methods show some localization capabilities, ours often predicts sharper and more clear boundaries, indicating a better understanding of the localization of objects. Row 4 shows the objects in chaotic environments, which is hard to recognize even with human eyes. Except for foreground objects, we also test the ability on the background stuff (row 5). It is interesting to see that even for background, ours can still distinguish it with foreground objects. 

%##################################################################################################
\begin{figure}[t]
\centering
\includegraphics[width=\linewidth]{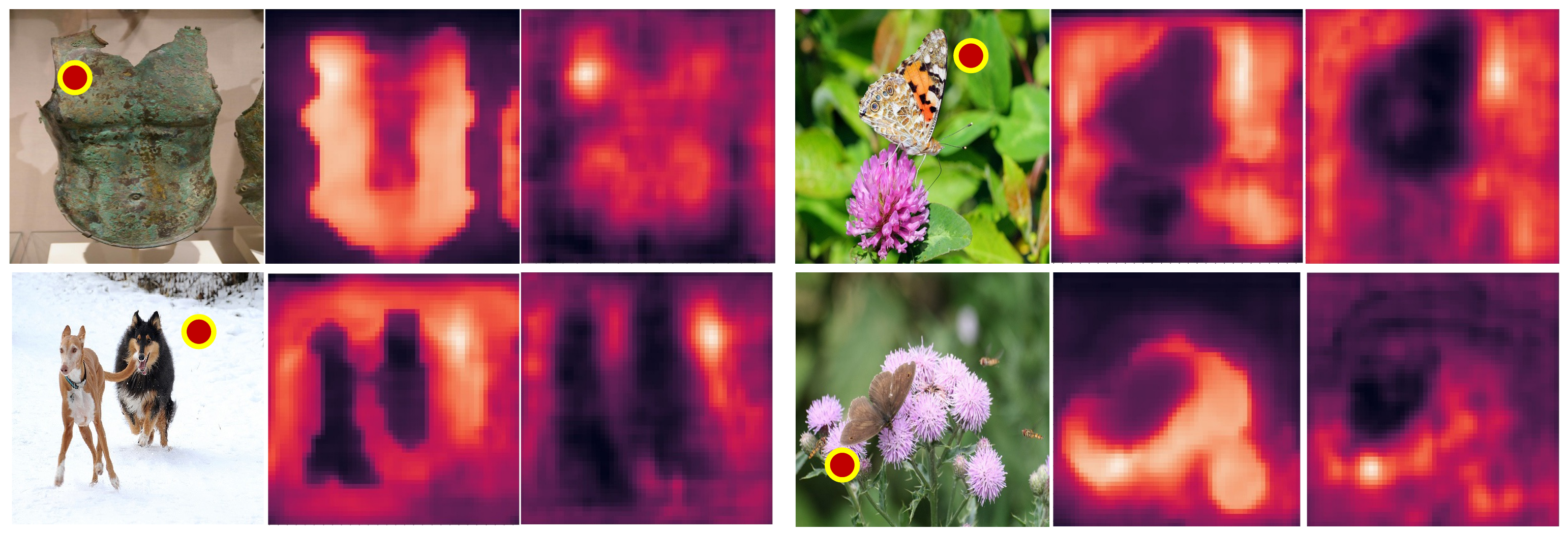}
\vspace{-1.5em}
\caption{{\bf Point affinity \emph{with} or \emph{without} affinity distillation.} In each of the 4 groups, we show (from left to right) the original image, ours \emph{with} point affinity distillation and ours \emph{without}. As can be seen, the distillation loss plays a key role in capturing object boundaries, as shown in the 4 groups of examples.
\label{fig:pointaffinityhelps}}
\vspace{-1em}
\end{figure}
%##################################################################################################

\paragraph{Failure cases.} 
We also give some failure cases from our model in \cref{fig:badcase}. On the left we show \emph{under-segmentation}, where a segment contains more objects than it should be. For example, in the first image, both the man and the running machine have higher similarity to the chosen point. On the contrary, on the right we show \emph{over-segmentation}, where a segment does not cover the entire object. For example, the face of the woman has higher similarity to the chosen point, while the clothes and wig have lower similarities -- ideally they should all belong to the same person. We believe this is reasonable in our unsupervised setting: without definition of object classes, the model can at best form groups using low-level cues such as textures or colors; therefore, it can miss semantic-level grouping of objects. 

%##################################################################################################
\begin{table}[t]
\centering
\small
\tablestyle{2pt}{1.1}
\begin{tabular}{x{42}x{42}|x{42}x{42}|x{43}}
random & supervised & image- & region- & ours \\ 
\shline
15.3 & 22.9 & 33.1 & 33.8 & \textbf{52.0} 
\end{tabular}
\vspace{-.5em}
\caption{{\bf Quantitative metric} to compare VOC visualizations from different pre-training methods. Our point-level region contrast outperforms all baselines ranging from random, supervised pre-training and self-supervised pre-training at various levels.
\label{tab:quant_point_affinity}}
\vspace{-1em}
\end{table}
%##################################################################################################

\paragraph{Affinity distillation helps localization.} We visualize our method with or without point affinity distillation in \cref{fig:pointaffinityhelps}. We find the distillation loss plays a key role in capturing object boundaries for better localization.

\paragraph{Quantitative Metric.}
We quantitatively evaluate the visualizations from different pre-training methods on VOC {\fontfamily{pcr}\selectfont val2007}. For each ground-truth object, we pick its center point and calculate the similarity from this point to the rest of the image with the pre-trained model to generate segmentation masks. We pick a threshold to keep 80\%
of the entire affinity map. The mask is then benchmarked with Jaccard similarity, defined as the intersection over union (IoU) between the predicted mask and the ground-truth one. 

Our baselines are: random (no pre-training), supervised (on ImageNet-1K), MoCo v2 (image-level) and region-level. The results are summarized in \cref{tab:quant_point_affinity}.
As expected, point-level region contrast significantly outperforms others.

%##################################################################################################
\begin{figure}[t]
\centering
\includegraphics[width=.9\linewidth]{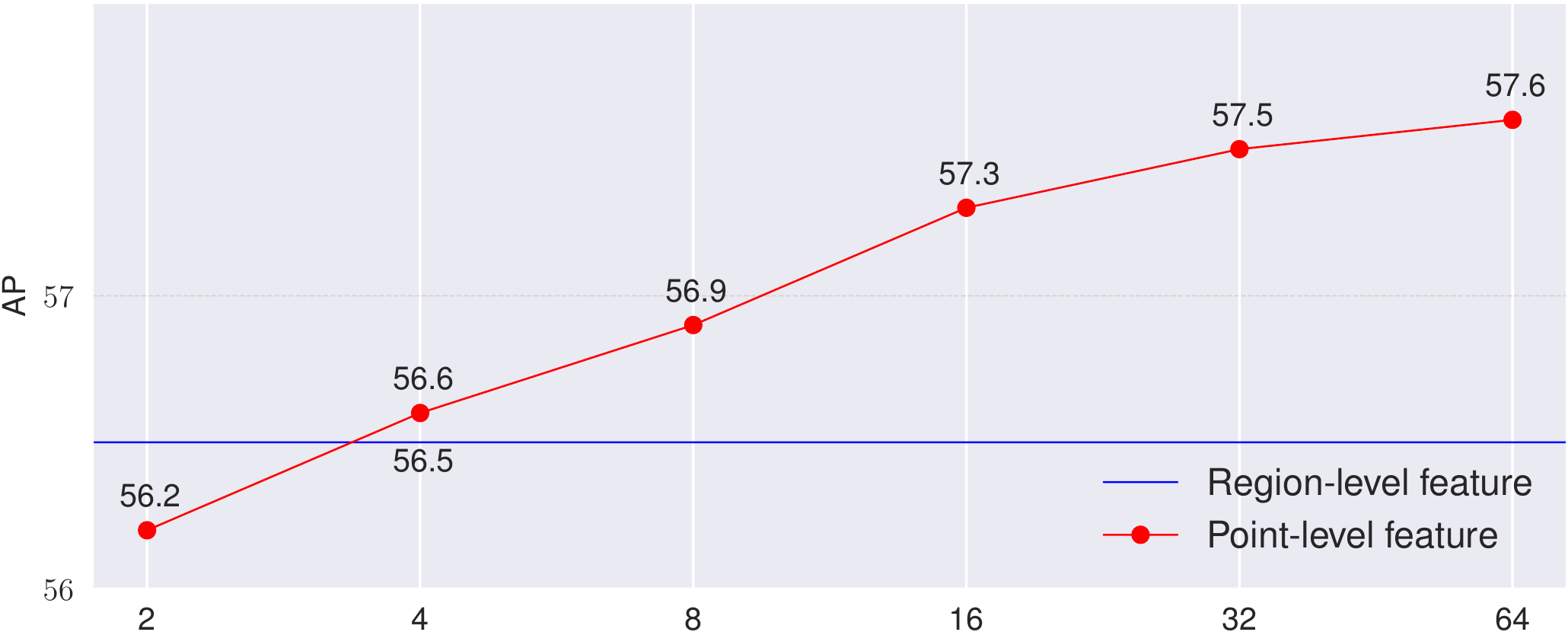}
\caption{{\bf Point-level \vs region-level features}. We check how many points are needed to match a region-level representation when pre-trained on ImageNet-1K. Along the horizontal axis the number of points increases from 2 to 64 for point-level features. The pre-trained representation can already match region-level features (blue line) in VOC AP with only 4 points.
\label{fig:pointvsreg}}
\vspace{-1em}
\end{figure}
%##################################################################################################

\subsection{Main Ablation Studies\label{sec:ablation}}
For our main ablation analysis, we begin with point-level contrastive learning in \cref{sec:ablaterobust}, showing its effectiveness to represent regions and robustness to inferior initial regions compared to a region-based counterpart. Then we discuss and compare possible point affinity distillation strategies in \cref{sec:ablatedistillation}. More ablations are found in the appendix. Throughout this section, we pre-train for 100 epochs on ImageNet-1K and 400 COCO epochs on COCO.

\subsubsection{Point-Level \vs Region-Level\label{sec:ablaterobust}}
We first design experiments to show the motivation and effectiveness of introducing point-level operations to region-level contrast. We conduct two experiments. 

First is to see how many points are needed to match the pooled region-level features. We pre-train on ImageNet for 100 epochs \emph{without point affinity loss} for fair comparisons, and report results with VOC object detection transfer. As shown in \cref{fig:pointvsreg}, we find with only \emph{4} points per-region, its AP (56.6) is already better than region-level contrast (56.5). Interestingly, more point-level features continue to benefit performance even up to 64 points, which suggests that the pooled, region-level features are not as effective as point-level ones for object detection pre-training.

\begin{figure}[t]
\centering
\hspace{-4mm}\includegraphics[width=.9\linewidth]{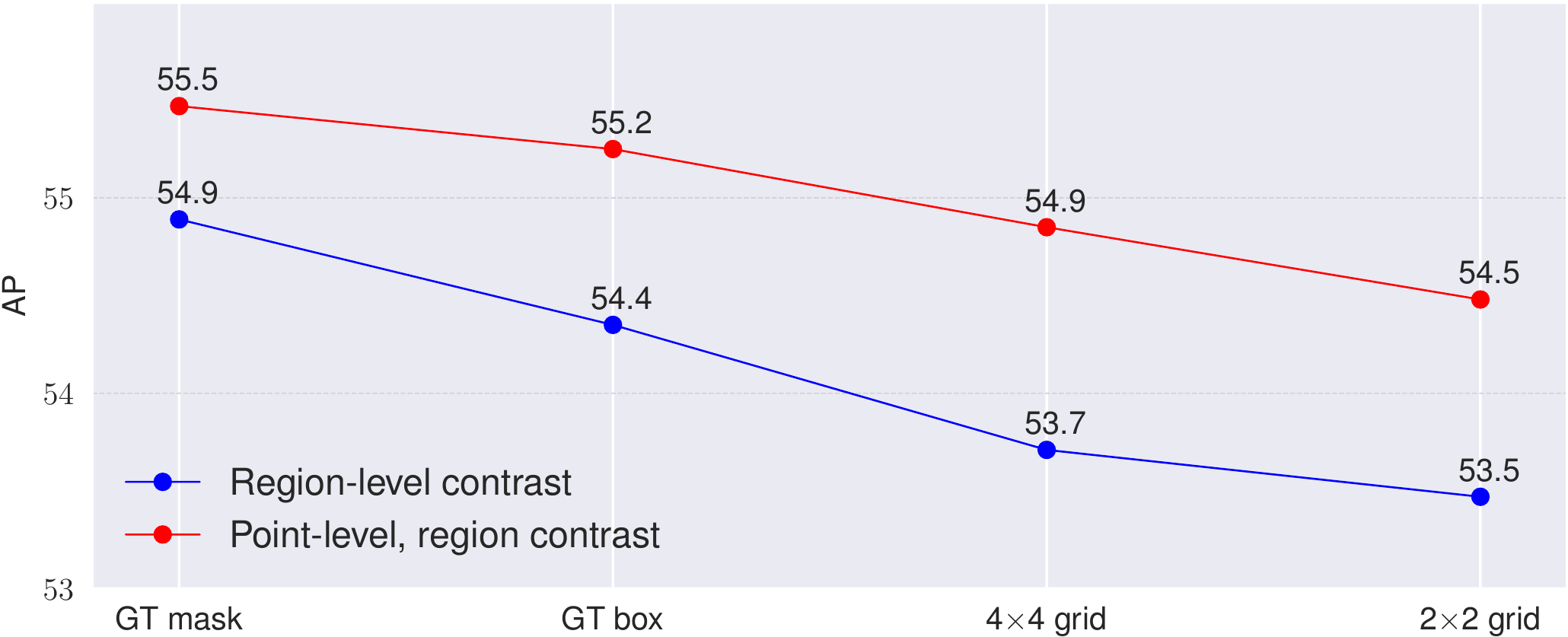}
\caption{{\bf Region quality \vs AP} comparison between our point-level region contrast (red) and region-level contrast with pooled-features, pre-trained on COCO. Along the horizontal axis the region quality degenerates: ground truth masks, ground truth bounding box, 4\x4 grid and 2\x2 grid. Our method is consistently better and is more resilient to the degeneration of region qualities.
\label{fig:robustness}}
\vspace{-1em}
\end{figure}

Second, we add back the point affinity loss and compare the robustness of our full method against contrast learning with aggregated region-level features~\cite{henaff2021efficient}. For this experiment, we pre-train on COCO as COCO is annotated with ground-truth object boxes/masks. $P{=}$16 points are used per-region and evaluation is also performed with VOC object detection. In \cref{fig:robustness} we gradually decrease the region quality, from highest (ground-truth mask), to lowest (2\x2 grid) with ground-truth box and 4\x4 grid in-between. Not only does point-level region contrast perform better than region-level contrast, the gap between the two \emph{increases} as the region quality degenerates from left to right. This confirms that our method is more \emph{robust} to initial region assignments and can work with all types of regions.

\subsubsection{Point Affinity Distillation Strategies\label{sec:ablatedistillation}}
For point affinity distillation, there are three possible strategies: 1) $\mA_{i'k'}$ as teacher (see \cref{eq:affinity} for its definition), $\mA_{ik'}$ as student (default); 2) $\mA_{ik'}$ as teacher, $\mA_{ik}$ as student; 3) $\mA_{i'k'}$ as teacher, $\mA_{ik}$ as student, which requires an \emph{extra} forward pass with momentum encoders.

Strategy 1) achieves 58.0 AP. Switching to strategy 2) slightly degenerates AP to 57.6, and strategy 3) yields the same AP as 1) while requiring extra computations. Therefore we set 1) as our default setting.

%##################################################################################################
%##################################################################################################

\section{Conclusion\label{sec:conclusion}} 
Balancing recognition and localization, we introduced point-level region contrast, which performs self-supervised pre-training by directly sampling individual point pairs from different regions. Compared to other contrastive formulations, our approach can learn both inter-image and intra-image distinctions, and is more resilient to imperfect unsupervised regions assignments. We empirically verified the effectiveness of our approach on multiple setups and showed strong results against state-of-the-art pre-training methods for object detection. We hope our explorations can provide new perspective and inspirations to the community.

\appendix

%##################################################################################################
\begin{table*}[!t]
\centering
%#################################################
% Point Ratio
%#################################################
\subfloat[
\textbf{Ratio $\alpha$ in \cref{eq:point}}
\label{tab:ablation_point_ratio_old}
]{%
\begin{minipage}{0.3\linewidth}
\begin{center}
\small
\tablestyle{5pt}{1.1}
\begin{tabular}{y{40}|x{20}x{20}x{20}}
$\alpha$ & AP & AP$_\text{50}$ & AP$_\text{75}$ \\ 
\shline
0 & 57.3 & 82.0 & 63.8 \\
0.3 & 57.5 & 82.1 & 64.1  \\
\baseline{0.5} & \baseline{\textbf{58.0}} & \baseline{\textbf{82.5 }}& \baseline{\textbf{64.7}} \\
0.7 & 57.6 & 82.3 & 64.3 \\
\end{tabular}
\end{center}
\end{minipage}
}
\hspace{1em}
%#################################################
% Image-Point Ratio
%#################################################
\subfloat[
\textbf{Ratio $\beta$ in \cref{eq:total_loss}}
\label{tab:ablation_image_point_ratio}
]{%
\begin{minipage}{0.3\linewidth}
\begin{center}
\small
\tablestyle{5pt}{1.1}
\begin{tabular}{y{40}|x{20}x{20}x{20}}
$\beta$ & AP & AP$_\text{50}$ & AP$_\text{75}$ \\ 
\shline
 0.3 & 56.4 & 81.5 & 62.4 \\
 0.5 & 57.1 & 82.1 & 63.8   \\
 \baseline{0.7 }& \baseline{\textbf{58.0}} & \baseline{\textbf{82.5}} & \baseline{\textbf{64.7}}  \\
 0.9 & 56.4 & 81.5 & 62.5 \\
\end{tabular}
\end{center}
\end{minipage}
}
\hspace{1em}
%#################################################
% Temperature of Knowledge Distillation
%#################################################
\subfloat[
\textbf{Distillation \emph{temp.}.} student ($\tau_s$), teacher ($\tau_t$)
\label{tab:temperature}
]{%
\begin{minipage}{0.3\linewidth}
\begin{center}
\small
\tablestyle{5pt}{1.1}
\begin{tabular}{y{16}|y{16}|x{20}x{20}x{20}}
$\tau_s$ & $\tau_t$ & AP & AP$_\text{50}$& AP$_\text{75}$ \\ 
\shline
0.1 & 0.04 & 57.7 & 82.3 & 64.4 \\
 \baseline{0.1}& \baseline{0.07} & \baseline{\textbf{58.0}} & \baseline{\textbf{82.5}} & \baseline{\textbf{64.7}} \\
0.2 & 0.15 & 57.3 & 82.1 & 64.2 \\
\multicolumn{5}{c}{}\\
\end{tabular}
\end{center}
\end{minipage}
}
\\
%#################################################
% Number of Points
%#################################################
\subfloat[
\textbf{Number of sampled points $P$}
\label{tab:ablation_point_number}
]{%
\begin{minipage}{0.3\linewidth}
\begin{center}
\small
\tablestyle{5pt}{1.1}
\begin{tabular}{y{40}|x{20}x{20}x{20}}
$P$ & AP & AP$_\text{50}$ & AP$_\text{75}$ \\ 
\shline
 8 & 57.1 & 81.8& 63.9 \\
 \baseline{16} & \baseline{58.0} & \baseline{82.5} & \baseline{64.7}  \\
 32 & \textbf{58.2} & \textbf{82.7} & \textbf{65.1} \\
\end{tabular}
\end{center}
\end{minipage}
}
\hspace{1em}
%#################################################
% Grid Number
%#################################################
\subfloat[
\textbf{Size of grid $n$}
\label{tab:ablation_grid_number}
]{%
\begin{minipage}{0.3\linewidth}
\begin{center}
\small
\tablestyle{5pt}{1.1}
\begin{tabular}{y{40}|x{20}x{20}x{20}}
$n$ & AP & AP$_\text{50}$ & AP$_\text{75}$ \\ 
\shline
 2 & 57.1 & 82.1 & 63.7 \\
 \baseline{4} & \baseline{\textbf{58.0}} & \baseline{\textbf{82.5}} & \baseline{\textbf{64.7}}  \\
 8 & 57.6 & 82.2 & 64.3 \\
\end{tabular}
\end{center}
\end{minipage}
}
\hspace{1em}
%#################################################
% Image-Point Ratio
%#################################################
\subfloat[
\textbf{Feature map resolution $R$}
\label{tab:ablation_resolution}
]{%
\begin{minipage}{0.3\linewidth}
\begin{center}
\small
\tablestyle{5pt}{1.1}
\begin{tabular}{y{40}|x{20}x{20}x{20}}
$R$ & AP & AP$_\text{50}$ & AP$_\text{75}$ \\ 
\shline
 14\x14 & 57.3 & 81.9 & 63.7  \\
 \baseline{56\x56} & \baseline{\textbf{58.0 }}& \baseline{\textbf{82.5}} & \baseline{\textbf{64.7}}  \\
 224\x224 & 57.2 & 82.2 & 63.6\\
\end{tabular}
\end{center}
\end{minipage}
}
%#################################################
\vspace{-.2em}
\caption{\textbf{Ablation studies.} For all of them, we pre-train our representation on ImageNet-1K for 100 epochs, and report the transfer results on VOC object detection. Our default settings are shown in gray. \label{tab:ablation_study}}
\vspace{-.5em}
\end{table*}
%##################################################################################################

\paragraph{Acknowledgements.} This work was partially supported by ONR N00014-21-1-2812. We thank the anonymous reviewers for their effort and valuable feedback to improve our work.

\section{More Ablation Analysis \label{sec:more_ablation}}

Beyond ablation analysis provided in the main paper (\cref{sec:ablation}), we provide three more groups of analysis in this appendix. They are: 1) balance between different losses during pre-training; 2) student and teacher temperatures in point affinity distillation; and 3) hyper-parameters for point sampling. Unless otherwise specified, we use ImageNet-1K and pre-train for 100 epochs. The results are reported on VOC object detection, all summarized in \cref{tab:ablation_study}.

\subsection{Balance Between Losses \label{sec:ablateratio}}

\paragraph{Contrastive \& affinity distillation.}
The hyper-parameter $\alpha$ in \cref{eq:point} serves as the weight to balance the two point based loss terms. By default we set $\alpha$ as 0.5 and we report the results of different $\alpha$ values in \cref{tab:ablation_point_ratio_old}. 

\paragraph{Image-level \& point-level.}
On top of point-level computation, we further leverage image-level loss. The hyper-parameter $\beta$ in~\cref{eq:total_loss} serves as the weight to balance the two loss terms.
We report the results of different $\beta$ values in \cref{tab:ablation_image_point_ratio}. We find when the image-level loss is small, the overall performance will be influenced, since the point-level task is harder to converge at the beginning. Adding image-level contrastive loss further enhances our method to balance localization and recognition capabilities.

\subsection{Temperatures in Point Affinity Distillation}
We now study the hyper-parameters for the student and teacher temperatures $\tau_s$ and $\tau_t$. Intuitively, we hope the output from the teacher is closer to a `one-hot' label~\cite{sohn2020fixmatch}, which means the teacher temperature is relatively smaller than the student one. We explored a few setups following this intuition, and summarize our observations in \cref{tab:temperature}.

From the default temperatures, our method is quite robust to the changes. For example, decreasing $\tau_t$ from 0.07 to 0.04 only slightly degenerates the performance. Varying both temperatures also do not affect much in the third row.

\subsection{Point Sampling\label{sec:ablatepoint}} 

\paragraph{Number of points $P$.}
For final loss which includes the point affinity, we also ablate the number of points. From the results in \cref{tab:ablation_point_number} we can observe the performance improves as the point number increases. We use point number $P{=}$16 as the default setting, where the performance starts to saturate. We report the results of different number of points. 

\paragraph{Number of grids $n$.}
In the default setting, the adopted grid number is 4\x4. We report the results of different number of grids in \cref{tab:ablation_grid_number}. From the table, we  observe the number of grid does not influence the results much.

\paragraph{Feature map resolution $R$.}
In the default setting, we up-sample the feature map to 56\x56. From \cref{tab:ablation_resolution}, we can observe that feature map resolution would also influence the results, with 56 x 56 further directing towards better AP.

\section{More Visualizations}
\begin{figure*}[!t]
\centering
\includegraphics[width=1\linewidth]{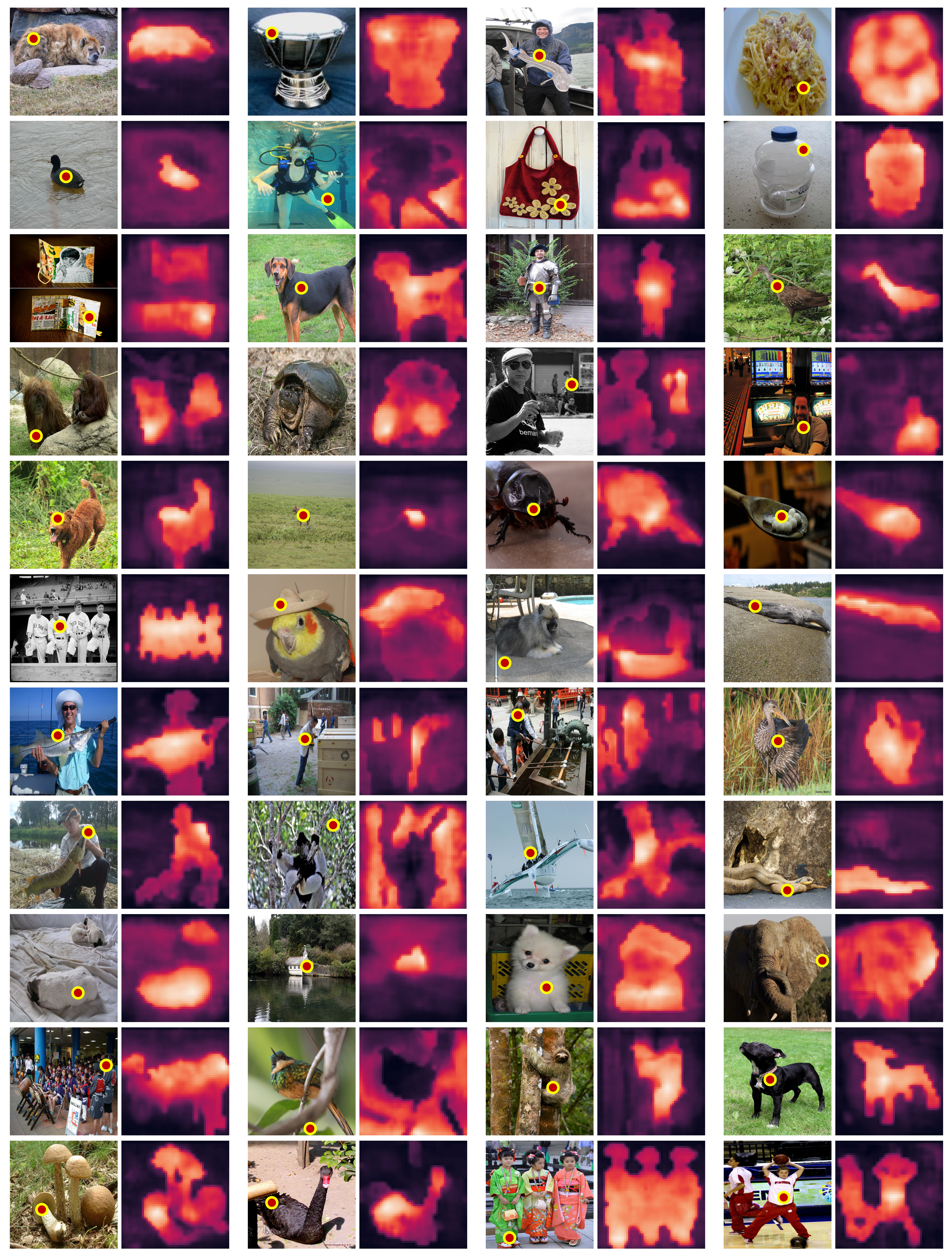}
\caption{\textbf{More visualizations} on ImageNet-1K with our point-level region contrast.\label{fig:sup_vis}}
\end{figure*}

We provide more visualizations with our method in \cref{fig:sup_vis}. We followed the same protocol which first pick a point (denoted by red circle), and then compute the affinity map with the pre-trained features. Brighter colors denote more similar points. Our method consistently generates masks with crisp boundaries.

\section{License of Assets}
%##################################################################################################
\begin{center}
\vspace{-.2em}
\small
\tablestyle{5pt}{1.1}
\begin{tabular}{y{40}|x{180}}
Dataset & Licence \\ 
\shline
ImageNet & https://image-net.org/download.php\\
COCO &  Creative Commons Attribution 4.0 License\\
Pascal VOC & http://host.robots.ox.ac.uk/pascal/VOC/ \\
Cityscapes & https://www.cityscapes-dataset.com/license/\\
\end{tabular}
\vspace{-.2em}
\end{center}
%##################################################################################################

%##################################################################################################
%##################################################################################################

\clearpage

%%%%%%%%% REFERENCES
{
\small
\bibliographystyle{ieee_fullname}
\bibliography{point}
}
\end{document}